\documentclass[11pt,twoside,fullpage]{article} 

\usepackage[round]{natbib}

\usepackage{url}

\usepackage{lineno}

\usepackage{graphicx}
\usepackage{amsmath}
\usepackage{url}
\usepackage[bookmarks=true%
,bookmarksnumbered=true%
,hypertexnames=false%
,colorlinks=true%
,linkcolor=black%
,citecolor=black%
,urlcolor=blue%
]{hyperref}

\usepackage{changepage}
\usepackage{booktabs}

\usepackage{multirow}
\usepackage{verbatim}
\usepackage{breqn}
\usepackage{enumitem}  

\usepackage{amsthm}
\usepackage{algorithm}
\usepackage{algorithmic}

\usepackage[text={15cm,20cm},centering]{geometry}

\title{Detecting Offensive Language in Tweets\\ Using Deep Learning}

\author{Georgios K. Pitsilis, Heri Ramampiaro and Helge Langseth\\
\small Department of Computer Science\\
\small Norwegian University of Science and Technology\\
\small NO-7491 Trondheim, Norway\\
\footnotesize {\tt \{georgios.pitsilis, heri.ramampiaro, helge.langseth\}@ntnu.no} \\ }

\date{}

\begin{document}

\maketitle

\begin{abstract}	
	This paper addresses the important problem of discerning hateful content in social media. We propose a detection scheme that is an ensemble of \emph{Recurrent Neural Network (RNN)} classifiers, and it incorporates various features associated with user-related information, such as the users' tendency towards \emph{racism} or \emph{sexism}.
	These data are fed as input to the above classifiers along with the word frequency vectors derived from the textual content. Our approach has been evaluated on a publicly available corpus of 16k tweets, and the results demonstrate its effectiveness in comparison to existing state of the art solutions. More specifically, 
	our scheme can successfully distinguish racism and sexism messages from normal text, and achieve higher classification quality than current state-of-the-art algorithms.\\

\end{abstract}


\section{Introduction}

Social media is a very popular way for people to express their opinions publicly and to interact with others online. In aggregation, social media can provide a reflection of public sentiment on various events.
Unfortunately, many users engaging online, either on social media, forums or blogs, will often have the risk of being targeted or harassed via abusive language, which may severely impact their online experience and the community in general. The existence of social networking services creates the need for detecting user-generated hateful messages prior to publication. All published text that is used to express hatred towards some particular group with the intention to humiliate its members is considered a hateful message.

Although hate speech is protected under the free speech provisions in the United States, there are other countries, such as Canada, France, United Kingdom, and Germany, where there are laws prohibiting it as being promoting violence or social disorder.
Social media services such as Facebook and Twitter have been criticized for not having done enough to prohibit the use of their services for attacking  people belonging to some specific race, minority etc. \citep{nytimes:2017}. 
They have announced though that they would seek to battle against racism and xenophobia \citep{dailymail}. 
Nevertheless, the current solutions deployed by them have attempted to address the problem with manual effort, relying on users to report offensive comments \citep{BBC}. 
This not only requires a huge effort by human annotators, but it also has the risk of applying discrimination under subjective judgment. 
Moreover, a non-automated task by human annotators would have strong impact on system response times, since a computer-based solution can accomplish this task much faster than humans.
The massive rise in the user-generated content in the above social media services, with manual filtering not being scalable, highlights the need for automating the process of on-line hate-speech detection.

Despite the fact that the majority of the solutions for automated detection of offensive text rely on Natural Language Processing (NLP) approaches, there is lately a tendency towards employing pure machine learning techniques like neural networks for that task. 
NLP approaches have the drawback of being complex, and to a large extent dependent on the language used in the text. This provides a strong motivation for employing alternative machine learning models for the classification task.
Moreover, the majority of the existing automated approaches depend on using pre-trained vectors (e.g. Glove, Word2Vec) as word embeddings to achieve good performance from the classification model. 
That makes the detection of hatred content unfeasible in cases where users have deliberately obfuscated their offensive terms with short slang words.

There is a plethora of unsupervised learning models in the existing literature  to deal with  hate-speech \citep{schmidt-wiegand:2017}, as well as in detecting the sentiment polarity in tweets \citep{Barnaghi2016}. At the same time, the supervised learning approaches have not been explored adequately so far.
While the task of sentence classification seems similar to that of sentiment analysis; nevertheless, in hate-speech even negative sentiment could still provide useful insight.
Our intuition is that the task of hate-speech detection can be further benefited by the incorporation of other sources of information to be used as features into a supervised learning model. 
A simple statistical analysis on an existing annotated dataset of tweets by  \cite{Waseem2016}, can easily reveal the existence of significant correlation between the user tendency in expressing opinions that belong to some offensive class (\emph{Racism} or \emph{Sexism}), and the annotation labels associated with that class. 
More precisely, the correlation coefficient value that describes such user tendency was found to be 0.71 for racism  in the above dataset, while that value reached as high as 0.76 for sexism.
In our opinion, utilizing  such user-oriented behavioural data for reinforcing an existing solution is feasible, because  such information is  retrieva2ble in real-world use-case scenarios like Twitter. 
This highlights the need to explore the user features more systematically to further improve the classification accuracy of a supervised learning system.

Our approach employs a neural network solution composed of multiple Long-Short-Term-Memory (LSTM) based classifiers, and utilizes user behavioral characteristics such as the tendency towards racism or sexism to boost performance.
Although our technique is not necessarily revolutionary in terms of the deep learning models used, we show in this paper that it is quite effective.

Our main contributions are: $i$) a deep learning architecture for text classification in terms of hateful content, which incorporates features derived form the users' behavioural data, $ii$) a language agnostic solution, due to no-use of pre-trained word embeddings, for detecting hate-speech, $iii$) an experimental evaluation of the model on a Twitter dataset, demonstrating the top performance achieved on the classification task.
Special focus is given to investigating how the additional features concerning the users' tendency to utter hate-speech, as expressed by their previous history, could leverage the performance.
To the best of our knowledge, there has not been done any previous study on exploring features related to the users tendency in hatred content that used a deep learning model. 

The rest of the paper is organized as follows.
In Section \ref{Statement} we describe the problem of hate speech in more detail, and we refer to the existing work in the field in Section \ref{relWork}. In Section \ref{Approach} we present our proposed model, while in Section \ref{Evaluation} we refer to the dataset used, the evaluation tests we performed and we discuss the results received. Finally, in Section \ref{Conclusion} we summarize our contributions and discuss the future work.

\section{Problem Statement}
\label{Statement}

The problem we address in this work can be formally described as follows:
Let $p$ be an unlabeled short sentence composed of a number of words, posted by a user $u$. 
Let $N$,$S$,$R$ be three classes denoting Neutrality, Sexism and Racism respectively in a textual content. Members of these classes are those postings with content classified as belonging to the corresponding class, for which the following holds: $N \cap S \cap R = \emptyset$.
Further, given that user $u$ has a previous history of message postings $P_u : p \notin P_u$, we assume that any previous posting $p_u \in P_u$ by that user is already labeled as belonging to any of the classes N,S,R.
Similarly, other postings by other users have also been labeled accordingly, forming up their previous history.
Based on these facts, the problem is to identify the class, which the unlabeled sentence $p$ by user $u$ belongs to.

The research question we address in this work is:

\begin{quote}\it
How to effectively identify the class of a new posting, given the identity of the posting user and the history of postings related to that user?
\end{quote}

To answer this question, our main goals can be summarized as follows:
\begin{itemize}
    \item To develop a novel method that can improve the state-of-art approaches within hate-speech classification, in terms of classification performance / accuracy.
    \item To investigate the impact of incorporating information about existing personalized labeled postings from users' past history on the classification performance / accuracy.
\end{itemize}

Note that existing solutions for automatic detection are still falling short to effectively detect abusive messages. Therefore there is a need for new algorithms which would do the job of classification of such content more effectively and efficiently. Our work is towards that direction.

\section{Related Work}
\label{relWork}

Simple word-based approaches, if used for blocking the posting of text or blacklisting users, not only fail to identify subtle offensive content, but they also affect the freedom of speech and expression. 
The word ambiguity problem -- that is, a word can have different meanings in different contexts -- is mainly responsible for the high false positive rate in such approaches. 
Ordinary NLP approaches on the other hand, are ineffective to detect unusual spelling, experienced in user-generated comment text. 
This is best known as the \textit{spelling variation} problem, and it is caused either by unintentional or intentional replacement of single characters in a token, aiming to obfuscate the detectors.

In general, the complexity of the natural language constructs renders the task quite challenging.
The employment of supervised learning classification methods for hate speech detection is not new. \cite{VignaCDPT17} reported performance for a simple LSTM classifier not better than an ordinary SVM, when evaluated on a small sample of Facebook data for only 2 classes (\emph{Hate, No-Hate}), and 3 different levels of strength of hatred.
\cite{DavidsonWMW17} described another way of detecting offensive language in tweets, based on some supervised model. They differentiate hate speech from offensive language, using a classifier that involves naive Bayes, decision trees and SVM.
Also, \cite{Nobata:2016} attempted to discern abusive content with a supervised model combining various linguistic and syntactic features in the text, considered at character uni-gram and bi-gram level, and tested on Amazon data. 
In general, we can point out the main weaknesses of NLP-based models in their non-language agnostic nature and the low scores in detection.

Unsupervised learning approaches are quite common for detecting offensive messages in text by applying concepts from NLP to exploit the lexical syntactic features of sentences  \citep{ChenZZX12}, or using AI-solutions and bag-of-words based text-representations \citep{Warner:2012}. 
The latter is known to be less effective for automatic detection, since hatred users apply various obfuscation tricks, such as replacing a single character in offensive words.
For instance, applying a binary classifier onto a \emph{paragraph2vec} representation of words has already been attempted on Amazon data in the past~\citep{Djuric:2015}, but it only performed well on a binary classification problem. 
Another unsupervised learning based solution is the work by \cite{Waseem2016b}, in which the authors proposed a set of criteria that a tweet should exhibit in order to be classified as offensive.
They also showed that differences in geographic distribution of users have only marginal effect on the detection performance.
Despite the above observation, we explore other features that might be possible to improve the detection accuracy in the solution outlined below.

The work by \cite{Waseem2016} applied a crowd-sourced solution to tackle hate-speech, with the creation of an additional dataset of annotations to extend the existing corpus. The impact of the experience of annotators in the classification performance was investigated.
The work by \cite{jha2017} dealt with the classification problem of tweets, but their interest was on sexism alone, which they distinguished into `Hostile', `Benevolent' or `Other'.
While the authors used the dataset of tweets from \cite{Waseem2016b}, they treated the existing \emph{`Sexism'} tweets as being of class \emph{`Hostile'}, while they collected their own tweets for the \emph{`Benevolent'} class, on which they finally applied the FastText by \cite{joulin2016bag}, and SVM classification.

\cite{Badjatiya:2017} approached the issue with a supervised learning model that is based on a neural network.
Their method achieved higher score over the same dataset of tweets than any unsupervised learning solution known so far. That solution uses an LSTM model, with features extracted by character \emph{n-grams}, and assisted by Gradient Boosted Decision Trees.
Convolution Neural Networks (CNN) has also been explored as a potential solution in the hate-speech problem in tweets, with character \emph{n-grams} and \emph{word2vec} pre-trained vectors being the main tools.
For example, \cite{Park2017} transformed the classification into a 2-step problem, where abusive text first is distinguished from the non-abusive, and then the class of abuse (\emph{Sexism} or \emph{Racism}) is determined. \cite{Gambck2017} employed pre-trained CNN vectors in an effort to predict four classes. They achieved slightly higher \emph{F-score} than  character \emph{n-grams}.

In spite of the high popularity of NLP approaches in hate-speech classification
\citep{schmidt-wiegand:2017}, we believe there is still a high potential for deep learning models to further contribute to the issue. 
At this point it is also relevant to note the inherent difficulty of the challenge itself, which can be clearly noted by the fact that no solution thus far has been able to obtain an F-score above 0.93.

\section{Description of our Recurrent Neural Network-based Approach}
\label{Approach}

The power of neural networks comes from their ability to find data representations that are useful for classification. 
Recurrent Neural Networks (RNN) are a special type of neural network, which can be thought of as the addition of loops to the architecture. RNNs use back propagation in the training process to update the network weights in every layer.
In our experimentation we used a powerful type of RNN known as \emph{Long Short-Term Memory Network} (LSTM).
Inspired by the work by \cite{Badjatiya:2017}, we experiment with combining various LSTM models enhanced with a number of novel features in an ensemble.
%
More specifically we introduce:

\begin{itemize}
\item A number of additional features concerned with the users' tendency towards hatred behaviour. 

\item An architecture, which combines the output by various LSTM classifiers to improve the classification ability. 
\end{itemize}

\subsection{Features}
\label{features}
We first elaborate into the details of the features derived to describe each user's tendency towards each class (\emph{Neutral}, \emph{Racism} or \emph{Sexism}), as captured in their tweeting history.
In total, we define the three features $t_{Na}$, $t_{Ra}$, $t_{Sa}$, representing a user's tendency towards posting \emph{Neutral}, \emph{Racist} and \emph{Sexist} content, respectively.
We let $m_a$ denote the set of tweets by user $a$, and use $m_{N,a}$, $m_{R,a}$ and $m_{S,a}$ to denote the subsets of those tweets that have been labeled as \emph{Neutral}, \emph{Racist} and \emph{Sexist} respectively. Now, the features are calculated as 
$t_{N,a} = |m_{N,a}|/|m_a|$,
$t_{R,a} = |m_{R,a}|/|m_a|$,and
$t_{S,a} = |m_{S,a}|/|m_a|$.

Furthermore, 
we choose to model the input tweets in the form of vectors using word-based frequency vectorization. 
That is, the words in the corpus are indexed based on their frequency of appearance in the corpus, and the index value of each word in a tweet is used as one of the vector elements to describe that tweet.
We note that this modelling choice provides us with a big advantage, because the model is independent of the language used for posting the message.

\subsection{Classification}

To improve classification ability we employ an ensemble of LSTM-based classifiers.

In total the scheme comprises a number of classifiers (3 or 5), each receiving the vectorized tweets 
together with behavioural features (see Section \ref{features})  as input.

\begin{figure}
\centering
\includegraphics[width=1.0\textwidth,angle=0]{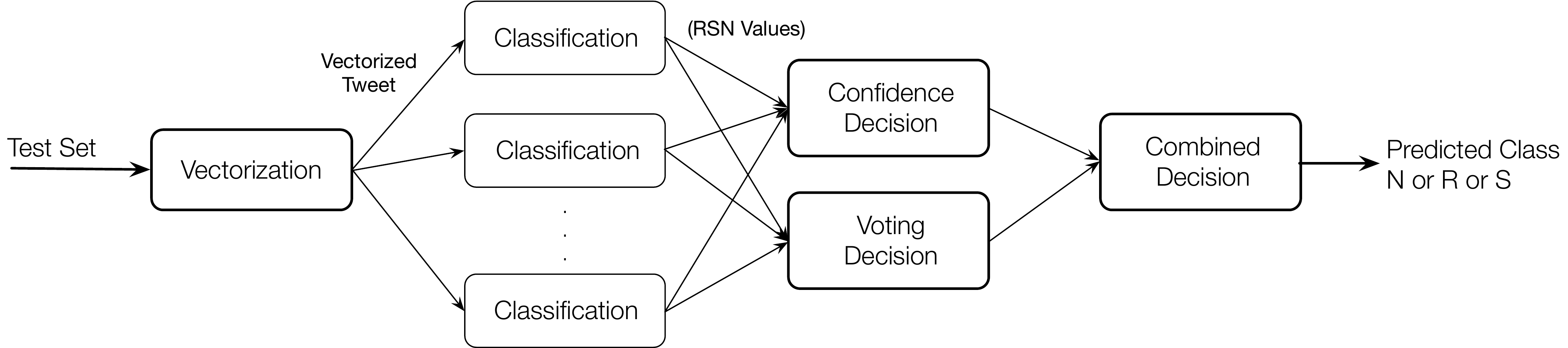}
\caption{High level view of the system with multiple classifiers}
\label{fig:high_level}
\end{figure}

The choice of various characteristics was done with the purpose to train the neural network with any data associations existing between the attributes for each tweet and the class label given to that tweet. In each case, the characteristic feature is attached to the already computed vectorized content for a tweet, thereby providing an input vector for one LSTM classifier.
A high level view of the architecture is shown in Figure~\ref{fig:high_level},
with the multiple classifiers.
The ensemble has two mechanisms for aggregating the classifications from the base classifiers; namely \emph{Voting} and \emph{Confidence}.  
The preferred method is majority voting, which is employed whenever at least two of the base classifiers agrees wrt.\ classification of a given tweet. When all classifiers disagree, the classifier with strongest confidence in its prediction is given preference. The conflict resolution logic is implemented in the \emph{Combined Decision} component.

\begin{algorithm}
\caption{Ensemble classifier}
\label{ensemble}
\begin{algorithmic}[1]
   \FOR{$tw \in \{ \text{tweets} \}$}
        \FOR{$cl \in \{$ classifiers $\}$}
          \STATE $(N_{cl},R_{cl},S_{cl}) \gets \text{classifier}_{cl}(tw)$
          \STATE $v_{cl} \gets \max(N_{cl},R_{cl},S_{cl})$
          \STATE $id_{cl} \gets \arg\max(N_{cl},R_{cl},S_{cl})$
        \ENDFOR
        \STATE $m \gets \text{mode}(id_1,id_2,id_3)$
        \IF{$m \in \{\text{Neutral, Racist, Sexism}\}$}
          \STATE decision $\gets m$
          \ELSE
          \STATE decision $\gets id_{\arg\max(v_1,v_2,v_3)}$
        \ENDIF
       \PRINT decision for $tw$
   \ENDFOR
\end{algorithmic}
\end{algorithm}

We present the above process in Algorithm~\ref{ensemble}. 
Here \textit{mode} denotes a function that provides the dominant value within the inputs classes $id_1,id_2,id_3$ and returns NIL if there is a tie, while \textit{classifier} is a function that returns the classification output in the form of a tuple ({\it N}eutral, {\it R}acism, {\it S}exism).

\section{Evaluation setup - Results}
\label{Evaluation}

\subsection{Data Preprocessing}

Before training the neural network with the labeled tweets, it is necessary to apply the proper tokenization to every tweet. 
In this way, the text corpus is split into word elements, taking white spaces and the various punctuation symbols used in the language into account. 
This was done using the Moses\footnote{http://www.statmt.org/moses/} package for machine translation.

We choose to limit the maximum size of each tweet to be considered during training to 30 words, and padded tweets of shorter size with zeros.
Next, tweets are converted into vectors using word-based frequency, as described in Section \ref{features}.
To feed the various classifiers in our evaluation, we attach the feature values onto every tweet vector.

In this work we experimented with various combinations of attached features $t_{N,a}$, $t_{R,a}$, and $t_{S,a}$ that express the user tendency. 
The details of each experiment, including the resulting size of each embedding can be found in Table~\ref{tab:Schemes}, with the latter denoted `input dimension' in the table.

\begin{table*}[!htbp]
   \begin{adjustwidth}{-1in}{-1in}
   \centering
   \hspace*{-0.4in}
   \begin{tabular}{lllc}
   \toprule
   \bf{Combination} & \bf{Additional features} & \bf{Features} & \bf{Input Dimension} \\
   \midrule
    O   & No additional features  & - & 30 \\
    NS  & Neutral \& Sexism & $t_{N,a}$, $t_{S,a}$ & 32 \\
    NR  & Neutral \& Racism & $t_{N,a}$, $t_{R,a}$ & 32 \\
    RS  & Racism \& Sexism & $t_{R,a}$, $t_{S,a}$ & 32 \\
    NRS & Neutral, Racism \& Sexism & $t_{N,a}$, $t_{R,a}$, $t_{S,a}$ & 33 \\
   \bottomrule
   \end{tabular}
   \end{adjustwidth}
   \caption{Combined features in proposed schemes}
   \label{tab:Schemes}

\end{table*}

\subsection{Deep learning model}

\begin{figure}
\centering
\hspace*{-0.27in}
\includegraphics[width=1\textwidth]{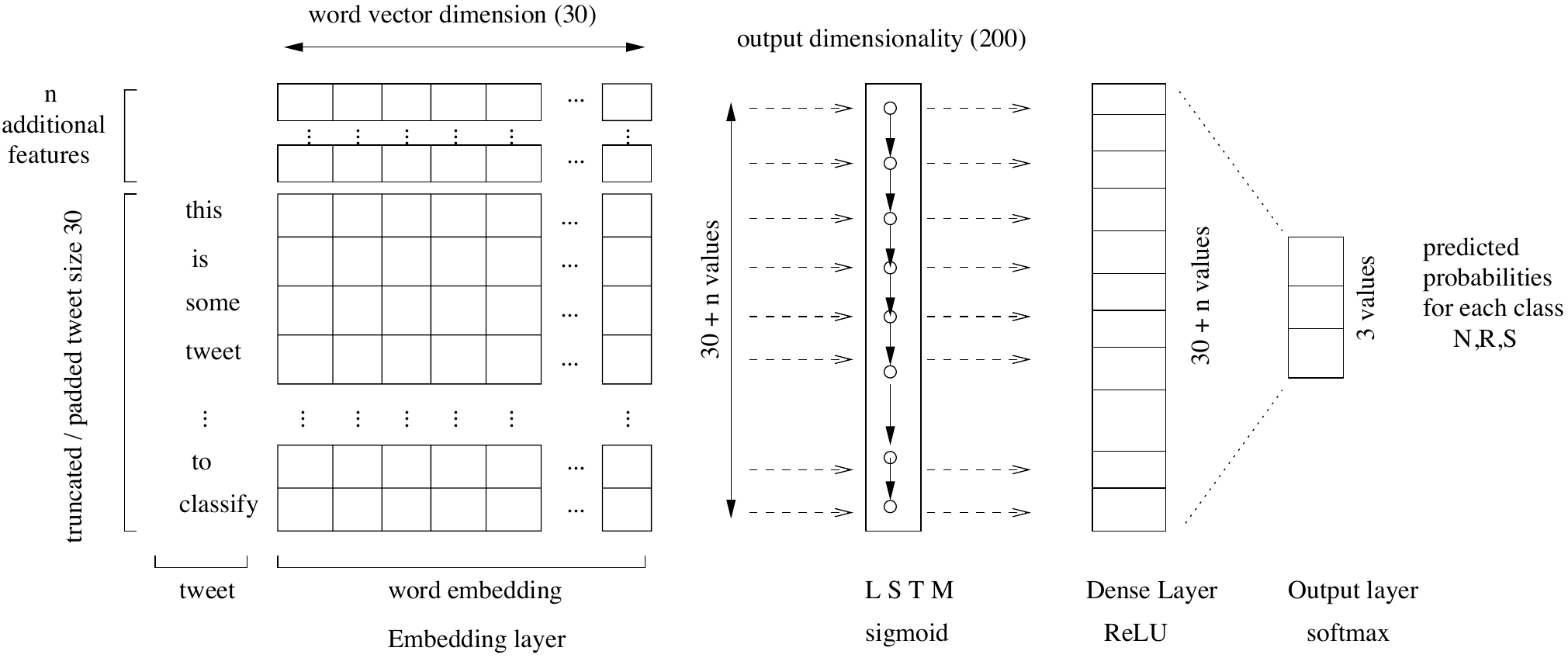}
\caption{Our deep learning model}
\label{fig:model}
\end{figure}

In our evaluation of the proposed scheme, each classifier is implemented as a deep learning model having four layers, as illustrated in Figure~\ref{fig:model}, and is described as follows:

\begin{itemize}
\item \emph{The Input (a.k.a Embedding) Layer}.
The input layer's size is defined by the number of inputs for that classifier. 
This number equals the size to the word vector plus the number of additional features.
The word vector dimension was set to 30 so that to be able to encode every word in the vocabulary used.

\item\emph{The hidden layer}. The sigmoid activation was selected for the the hidden LSTM layer. Based on preliminary experiments the dimensionality of the output space for this layer was set to 200. 
This layer is fully connected to both the Input and the subsequent layer.
\item \emph{The dense layer}. The output of the LSTM was run through an additional layer to improve the learning and obtain more stable output. The \emph{ReLU} 
activation function was used. Its size was selected equal to the size of the input layer.
\item \emph{The output layer}. This layer has 3 neurons to provide output in the form of probabilities for each of the three classes \emph{Neutral}, \emph{Racism}, and \emph{Sexism}.
The softmax activation function was used for this layer.
\end{itemize}

In total we experimented with 11 different setups of the proposed scheme, each with a different ensemble of classifiers, see Table~\ref{tab:Ensembles}.

\begin{table*}[!htbp]
\small
   \centering
   \hspace*{-0.4in}
   \begin{tabular}{|c|c|c|c|c|c|}
   \hline
   \bf{Tested} & \bf{Classifier} & \bf{Classifier} & \bf{Classifier} & \bf{Classifier} & \bf{Classifier}\\
   \bf{Scheme} &  \bf{I} & \bf{II} & \bf{III}  & \bf{IV}  & \bf{V}\\
   \hline
   \hline
   \emph{(i)}   & O   & NRS  & NR  & - & - \\
   \hline
   \emph{(ii)}  & O   & NRS  & NS  & - & - \\
   \hline
   \emph{(iii)} & O   & NRS  & RS  & - & - \\
   \hline
   \emph{(iv)}  & O   & NS   & RS  & - & - \\
   \hline   
   \emph{(v)}   & O   & NS   & NR  & - & - \\ 
   \hline   
   \emph{(vi)}  & O   & RS   & NR & - & - \\
   \hline   
   \emph{(vii)} & NRS & NR   & RS & - & - \\
   \hline
   \emph{(viii)} & NRS & NR  & NS & - & - \\
   \hline
   \emph{(ix)}  & NRS & NS  & RS  & - & - \\
   \hline
   \emph{(x)}   & NS & RS  & NR & - & - \\
   \hline
   \emph{(xi)}  & O & NS & RS  & NR & NRS \\
   \hline
   \end{tabular}
   \caption{Evaluated ensemble schemes}
   \label{tab:Ensembles}
\end{table*}

\subsection{Dataset}
We experimented with a dataset of approximately 16k short messages from Twitter, that was made available by \cite{Waseem2016b}. 
The dataset contains 1943 tweets labeled as \emph{Racism}, 3166 tweets labeled as \emph{Sexism} and 10889 tweets labeled as \emph{Neutral} (i.e., tweets that neither contain sexism nor racism). 
There is also a number of dual labeled tweets in the dataset.
More particularly, we found 42 tweets labeled both as both `Neutral' and `Sexism', while six tweets were labelled as both `Racism' and `Neutral'.
According to the dataset providers, the labeling was performed manually.\footnote{The small discrepancy observed in the class quantities with regard to those mentioned in the original dataset is due to fact that, at the time we performed the evaluation, a number of tweets were not retrievable.}

The relatively small number of tweets in the dataset makes the task more challenging.
As reported by several authors already, the dataset is imbalanced, with a majority of neutral tweets. 
Additionally, we used the public Twitter API to retrieve additional data associated with the user identity for each tweet in the original dataset. 

\subsection{Experimental Setting}

To produce results in a setup comparable with the current state of the art   \citep{Badjatiya:2017}, we performed 10-fold cross validation and calculated the \emph{Precision},\emph{Recall} and \emph{F-Score} for every evaluated scheme.
We randomly split each training fold into 15\% validation and 85\% training, while  performance is evaluated over the remaining fold of unseen data.
The model was implemented using Keras\footnote{https://github.com/fchollet/keras}. We used \emph{categorical cross-entropy} as learning objective, and selected the {ADAM}  optimization algorithm \citep{KingmaB14}. 
Furthermore, the vocabulary size was set to 25000, and the batch-size during training was set to 500.

To avoid over-fitting, the model training was allowed to run for a maximum number of 100 epochs, out of which the optimally trained state was chosen for the model evaluation. 
An optimal epoch was identified so, such that the validation accuracy was maximized, while at the same time the error remained within $\pm 1\%$ of the lowest ever figure within the current fold.
Throughout the experiment we observed that the optimal epochs typically occurred after between the 30 and 40 epochs.

To achieve stability in the results produced, we ran every single classifier for 15 times and the output values were aggregated. In addition, the output from each single classifier run was combined with the output from another two single classifiers to build the input of an ensemble, producing $15^3$ combinations.
For the case of the ensemble that incorporates all five classifiers we restricted to using the input by only the first five runs of the single classifiers ($5^5$ combinations). That was due to the prohibitively very large number of combinations that were required.

\subsection{Results}
\begin{figure}[!ht]
   \label{fig:graph}
   
    \centering
   
    \vspace*{-1.3in}
    \hspace*{-0.45in}    
    \includegraphics[width=1.18\textwidth]{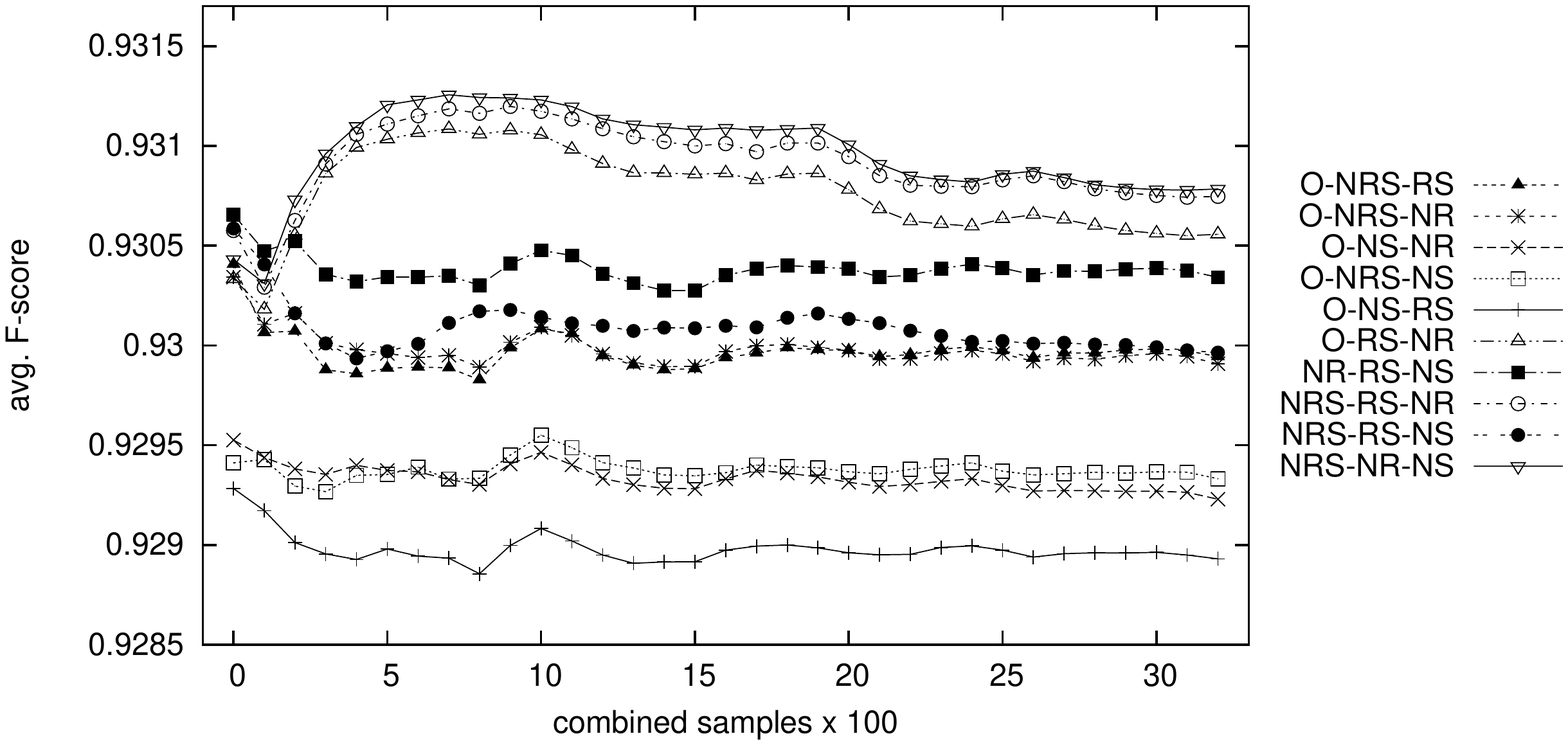}
    \vspace*{-0.3in}
    \caption{Aggregated value for F-score vs the number of experiment runs}
    \label{fig:progress}
\end{figure}

We now present the most interesting results from our experiments. 
For the evaluation we used standard metrics for classification accuracy, suitable for studying problems such as sentiment analysis. 
In particular we used \emph{Precision} and \emph{Recall}, with the former calculated as the ratio of the number of tweets correctly classified to a given class over the total number  of tweets classified to that class, while the latter  measures the ratio of messages correctly classified to a given class over the number of messages from that class.
Additionally, the \emph{F-score} is the harmonic mean of precision and recall, expressed as $ F = \frac{2 \cdot P \cdot R}{P + R}$.
For our particular case with three classes, \emph{P}, \emph{R} and \emph{F} are computed for each class separately, with the final \emph{F} value derived as the weighted mean of the separate $F$-scores:  $F=\frac{F_N \cdot N + F_R \cdot R + F_S \cdot S}{N+R+S}$; recall that $N=10889$, $S=3166$ and $R=1943$.
The results are shown in Table~\ref{tab:Results}, along with the reported results from state of the art approaches proposed by other researchers in the field. Note that the performance numbers \emph{P},\emph{R} and \emph{F} of the other state of the art approaches are based on the authors' reported data in the cited works.
Additionally, we report the performance of each individual LSTM classifier as if used alone over the same data (that is, without the ensemble logic).
The \emph{F-score} for our proposed approaches shown in the last column, is the weighted average value over the 3 classes (Neutral,Sexism,Racism). 
Moreover, all the reported values are average values produced for a number of runs of the same tested scheme over the same data. 
Figure~\ref{fig:progress} shows the \emph{F-Score} as a function of the number of training samples for each ensemble of classifiers. We clearly see that the models converge. 
For the final run the \textit{F-score} has standard deviation value not larger than 0.001, 
for all classifiers.

\begin{table*}[!htbp]
\small
   \begin{adjustwidth}{-1in}{-1in}
   \centering
   \hspace*{-0.4in}
   \begin{tabular}{|c|c|c|c|c|}
   \hline
   \bf{Approach} & \bf{Characteristics} & \bf{Precision} & \bf{Recall} & \bf{F-Score} \\
   \hline
   \hline
    single classifier \emph{(i)}    & O             & 0.9175 & 0.9218 & 0.9196 \\
    \hline
    single classifier \emph{(ii)}   & NS            & 0.9246 & 0.9273 & 0.9260 \\
    \hline
    single classifier \emph{(iii)}  & NR            & 0.9232 & 0.9259 & 0.9245 \\
    \hline
    single classifier \emph{(iv)}   & RS            & 0.9232 & 0.9264 & 0.9248 \\ 
    \hline
    single classifier \emph{(v)}    & NRS           & 0.9252 & 0.9278 & 0.9265 \\
   \hline
   \hline
   ensemble \emph{(i)}   & O + NRS + NR  & 0.9283   & 0.9315  & 0.9298  \\
   \hline
   ensemble \emph{(ii)}  & O + NRS + NS  & 0.9288   & 0.9319  & 0.9303  \\
   \hline
   ensemble \emph{(iii)} & O + NRS + RS  & 0.9283   & 0.9315  & 0.9299 \\
   \hline
   ensemble \emph{(iv)}  & O + NS + RS   & 0.9277   & 0.9310  & 0.9293 \\
   \hline
   ensemble \emph{(v)}   & O + NS + NR   & 0.9276   & 0.9308  & 0.9292 \\
   \hline
   ensemble \emph{(vi)}  & O + RS + NR   &  0.9273  & 0.9306  & 0.9290 \\
   \hline
   ensemble \emph{(vii)} & NRS + NR + RS & 0.9292   & 0.9319  &  0.9306 \\
   \hline
   ensemble \emph{(viii)} & NRS + NR + NS & 0.9295   &  0.9321 & 0.9308 \\
   \hline
   ensemble \emph{(ix)}  & NRS + NS + RS & 0.9294   & 0.9321  &  0.9308 \\
   \hline
   ensemble \emph{(x)}   & NS  + RS + NR & 0.9286   & 0.9314  & 0.9300 \\
   \hline
   ensemble \emph{(xi)}   & O + NS + RS + NR + NRS & 0.9305   & 0.9334  & \bf{0.9320} \\
   \hline
   \hline
   \cite{Badjatiya:2017} & LSTM + Random Embedding & 0.9300 & 0.9300 & 0.9300 \\
                         & + GBDT                  &        &        &        \\
   \hline
   \cite{Waseem2016b} & Unsupervised & 0.7290 & 0.7774 & 0.7391 \\
                      &     List of Criteria  & & & \\
   \hline
   \cite{Waseem2016} & Unsupervised  & 0.9159 & 0.9292 & 0.9153 \\
                     &     Expert annotators only             & & & \\
   \hline
   \cite{Park2017}  & 2 step HybridCNN           & 0.8270 & 0.8270 & 0.8270 \\ 
                    &    (Word Vec. / Char Vec.)  &       &        & \\
   \hline
   \end{tabular}
   \end{adjustwidth}
   \caption{Evaluation Results}
   \label{tab:Results}

\end{table*}

As can be seen in Table~\ref{tab:Results}, the work by \cite{Waseem2016b}, in which character \emph{n-grams} and gender information were used as features, obtained the quite low \emph{F-score} of $0.7391$.
Later work by the same author \citep{Waseem2016} investigated the impact of the experience of the annotator in the performance, but still obtaining a lower \emph{F-score} than ours. 
Furthermore, while the first part of the two step classification \citep{Park2017} performs quite well (reported an \emph{F-score} of 0.9520), it falls short in detecting the particular class the abusive text belongs to.
Finally, we observe that applying a simple LSTM classification with no use of additional features (denoted  `single classifier (i)' in Table~\ref{tab:Results}), achieves an \emph{F-score} that is below 0.93, something that is in line with other researchers in the field, see \cite{Badjatiya:2017}.

Very interestingly, the incorporation of features related to user's behaviour into the classification has provided a significant increase in the performance vs.\ using the textual content alone, $F=0.9295$ vs.\ $F=0.9089$.

Another interesting finding is the observed performance improvement by using an ensemble instead of a single classifier; some ensembles outperform the best single classifier.
Furthermore, the NRS classifier, which produces the best score in relation to other single classifiers, is the one included in the best performing ensemble.

In comparison to the approach by \cite{jha2017}, which focuses on various classes of Sexism, the results show that our deep learning model is doing better as far as detecting Sexism in general, outperforming the FastText algorithm they include in their experiments (F=0.87).
The inferiority of FastText over LSTM is also reported in the work by \cite{Badjatiya:2017}, as well as being inferior over CNN in, \cite{Park2017}.
In general, through our ensemble schemes is confirmed that deep learning can outperform any NLP-based approaches known so far in the task of abusive language detection.

We also present the performance of each of the tested models per class label in Table~\ref{tab:Detailed}.
Results by other researchers have not been included, as these figures are not reported in the existing literature.
As can be seen, \emph{sexism} is quite easy to classify in hate-speech, while  \emph{racism} seems to be harder; similar results were reported by \cite{DavidsonWMW17}. 
This result is consistent across all ensembles.

\begin{table*}
   \centering
    \begin{tabular}{|c|c|c|c|c|}
   \hline
   \bf{Proposed} &       &           &        &         \\ 
   \bf{Approach} & \bf{Class} & \bf{Precision} & \bf{Recall} & \bf{F-Score} \\
   \hline
   \hline
    \multirow{3}{*}{ensemble \emph{(viii)}} 
           & \emph{Neutral} & 0.9409 & 0.9609 & \bf{0.9508} \\
           & \emph{Racism} & 0.7522 & 0.6646 & \bf{0.7057} \\
           & \emph{Sexism} & 0.9991 & 0.9972 & \bf{0.9981} \\ 
   \hline
     \multirow{3}{*}{ensemble \emph{(ix)}} 
           & \emph{Neutral} & 0.9407 & 0.9612 & 0.9508 \\
           & \emph{Racism} & 0.7533 & 0.6627 & 0.7051 \\
           & \emph{Sexism} & 0.9986 & 0.9972 & 0.9979 \\    
   \hline
        \multirow{3}{*}{ensemble \emph{(vii)}} 
           & \emph{Neutral} & 0.9405 & 0.9611 & 0.9507 \\
           & \emph{Racism} & 0.7522 & 0.6616 & 0.7040 \\
           & \emph{Sexism} & 0.9990 & 0.9975 & 0.9983 \\    
   \hline
        \multirow{3}{*}{ensemble \emph{(xi)}} 
        
           & \emph{Neutral} & 0.9406 & 0.9631 & \bf{0.9517} \\
           & \emph{Racism}  & 0.7623 & 0.6617 & \bf{0.7084} \\
           & \emph{Sexism}  & 0.9992 & 0.9980 & \bf{0.9986} \\    
   
   \hline
   \end{tabular}
     \caption{Detailed Results for every Class Label}
   \label{tab:Detailed}   

\end{table*}

For completion, the confusion matrices of the best performing approach that employs 3 classifiers (ensemble \emph{viii}) as well as of the ensemble of the 5 classifiers (xi), are provided in Table~\ref{tab:Confusion}. The presented values is the sum over multiple runs.

The code and results associated with this paper will be available on-line soon at:
 \url{https://github.com/gpitsilis/hate-speech/}

\begin{table*}
\small
   \centering
   \begin{tabular}{|c|c|c|c|c|c|}
   \hline
   ensemble \emph{(viii)}
   & \multicolumn{5}{|c|} { \bfseries{\emph{Predicted Label}}  }  \\ 
   \hline
   \multirow{5}{*}{\bfseries{\emph{True Label}} } 
   
      &    & \bf{Racism} & \bf{Sexism} & \bf{Neutral} & sum \\
               \cline{2-6} 
               \cline{2-6} 
     & \bf{Racism} & 10655320 & 5635   & 24295   & 10685250  \\
     \cline{2-6} 
     & \bf{Sexism}  & 3943   & 4357971 & 2195711     & 6557625  \\
     \cline{2-6} 
     & \bf{Neutral}  & 5929   & 1430030 & 35314416 & 36750375 \\
     \cline{2-6} 
     \cline{2-6} 
     &   sum       & 10665192  & 5793636 & 37534422 & 53993250  \\
     &             &        &       &        & ($15998 \times 15^3$) \\
   \hline
   \hline
   ensemble \emph{(xi)}
   & \multicolumn{5}{|c|} { \bfseries{\emph{Predicted Label}}  }  \\ 
   \hline
   \multirow{5}{*}{\bfseries{\emph{True Label}} } 
   
      &    & \bf{Racism} & \bf{Sexism} & \bf{Neutral} & sum \\
               \cline{2-6} 
               \cline{2-6} 
     & \bf{Racism} & 9873754 &   991   & 19005   & 9893750  \\
     \cline{2-6} 
     & \bf{Sexism}  & 3034   & 4017687 & 2051154   & 6071875  \\
     \cline{2-6} 
     & \bf{Neutral}  & 4446   & 1252093 & 32771586 & 34028125 \\
     \cline{2-6} 
     \cline{2-6} 
     &   sum       & 9881234  & 5270771 & 34841745 & 49993750  \\
     &             &        &       &        & ($15998 \times 5^5$) \\
   \hline
   
   \end{tabular}
      \caption{Confusion Matrices of Results for the best performing approaches with 3 and 5 classifiers.}
   \label{tab:Confusion}   

\end{table*}

\section{Conclusions and Future Work}
\label{Conclusion}
In this work we present an ensemble classifier that is detecting hate-speech in short text, such as tweets. The input to the base-classifiers consists of not only the standard word uni-grams, but also a set of features describing each user's historical tendency to post abusive messages. 
Our main innovations are: \emph{i)} a deep learning architecture that uses word frequency vectorisation for implementing the above features, \emph{ii)} an experimental evaluation of the above model on a public dataset of labeled tweets, 
\emph{iii)} an open-sourced implementation built on top of Keras.

The results show that our approach outperforms the current state of the art, and
to the best of our knowledge, no other model has achieved better performance in classifying short messages.
The approach does not rely on pre-trained vectors, which provides a serious advantage when dealing with short messages of this kind. More specifically, users will often prefer to obfuscate their offensive terms using shorter slang words or create new words by `inventive' spelling and word concatenation.
For instance, the word `Islamolunatic' is not available in the popular pre-trained word embeddings (Word2Vec or GloVe), even though it appears with a rather high frequency in racist postings. Hence,  word frequency vectorization is preferable to the  pre-trained word embeddings used in prior works if one aims to build a language-agnostic solution.

We believe that deep learning models have a high potential wrt.\  classifying text or analyzing the sentiment in general. In our opinion there is still space for further improving the classification algorithms.

For future work we plan to investigate other sources of information that can be utilized 
to detect hateful messages. In addition,  we intend to generalize the output received in the current experiment, with evaluation over other datasets, including analyzing texts written in different languages.

\bibliographystyle{plainnat}
\bibliography{bibliography}

\end{document}